\documentclass[11pt]{article}
\pdfoutput=1
\usepackage{acl2015}
\usepackage{times}
\usepackage{url}
\usepackage{url}
\usepackage{color}
\usepackage{float}
\usepackage{tabularx}
\usepackage{subfig}
\usepackage{multirow}
\usepackage{latexsym}
\usepackage{amsmath}
\usepackage{amssymb}
\usepackage{multirow}
\usepackage[lined,boxed,commentsnumbered]{algorithm2e}
\usepackage{tikz-dependency}
\usepackage{pgfplots}
\pgfplotsset{compat=1.5}

\DeclareMathOperator*{\argmax}{arg\,max}
\def\x{\mathbf{x}}

\def\w{\mathbf{w}}
\def\W{\mathbf{W}}
\def\bd{\mathbf{d}}
\def\z{\mathbf{z}}
\def\bv{\mathbf{v}}
\def\p{\mathbf{p}}

\title{A Re-ranking Model for Dependency Parser \\with Recursive Convolutional Neural Network}

\author{Chenxi Zhu, Xipeng Qiu\thanks{\quad Corresponding author.}, Xinchi Chen, Xuanjing Huang\\
  Shanghai Key Laboratory of Intelligent Information Processing, Fudan University\\
School of Computer Science, Fudan University\\
825 Zhangheng Road, Shanghai, China\\
\{czhu13,xpqiu,xinchichen13,xjhuang\}@fudan.edu.cn}

\begin{document}
\maketitle

\begin{abstract}
In this work, we address the problem to model all the nodes (words or phrases) in a dependency tree with the dense representations.
We propose a recursive convolutional neural network (RCNN) architecture to capture syntactic and compositional-semantic representations of phrases and words in a dependency tree. Different with the original recursive neural network, we introduce the convolution and pooling layers, which can model a variety of compositions by the feature maps and choose the most informative compositions by the pooling layers.
Based on RCNN, we use a discriminative model to re-rank a $k$-best list of candidate dependency parsing trees. The experiments show that RCNN is very effective to improve the state-of-the-art dependency parsing on both English and Chinese datasets.
\end{abstract}

\section{Introduction}


Feature-based discriminative supervised models have
achieved much progress in dependency parsing \cite{Nivre:2004,Yamada:2003,McDonald:2005}, which typically use millions of discrete binary features generated from a limited size training data. However, the ability of these models is restricted by the design of features. The number of features could be so large that the result models are too complicated for practical use and prone to overfit on training corpus due to data sparseness.

Recently, many methods are proposed to learn various distributed representations on both syntax and semantics levels. These distributed representations have been extensively applied on many natural language processing (NLP) tasks, such as syntax \cite{Turian:2010,mikolov2010recurrent,collobert2011natural,Chen:2014a} and semantics \cite{Huang:2012,Mikolov:2013}.
Distributed representations are to represent words (or phrase) by the dense, low-dimensional and real-valued vectors, which help address the curse of dimensionality and have better generalization than discrete representations.

For dependency parsing, \newcite{Chen:2014} and \newcite{Bansal:2014} used the dense vectors (embeddings) to represent words or features and found these representations are complementary to the traditional discrete feature representation.
However, these two methods only focus on the dense representations (embeddings) of words or features. These embeddings are pre-trained and keep unchanged in the training phase of parsing model, which cannot be optimized for the specific tasks.

Besides, it is also important to represent the (unseen) phrases with dense vector in dependency parsing. Since the dependency tree is also in recursive structure, it is intuitive to use the recursive neural network (RNN), which is used for constituent parsing \cite{Socher2013parsing}. However, recursive neural network can only process the binary combination and is not suitable for dependency parsing, since a parent node may have two or more child nodes in dependency tree.

\begin{figure}
  \centering
  \includegraphics[width=0.48\textwidth]{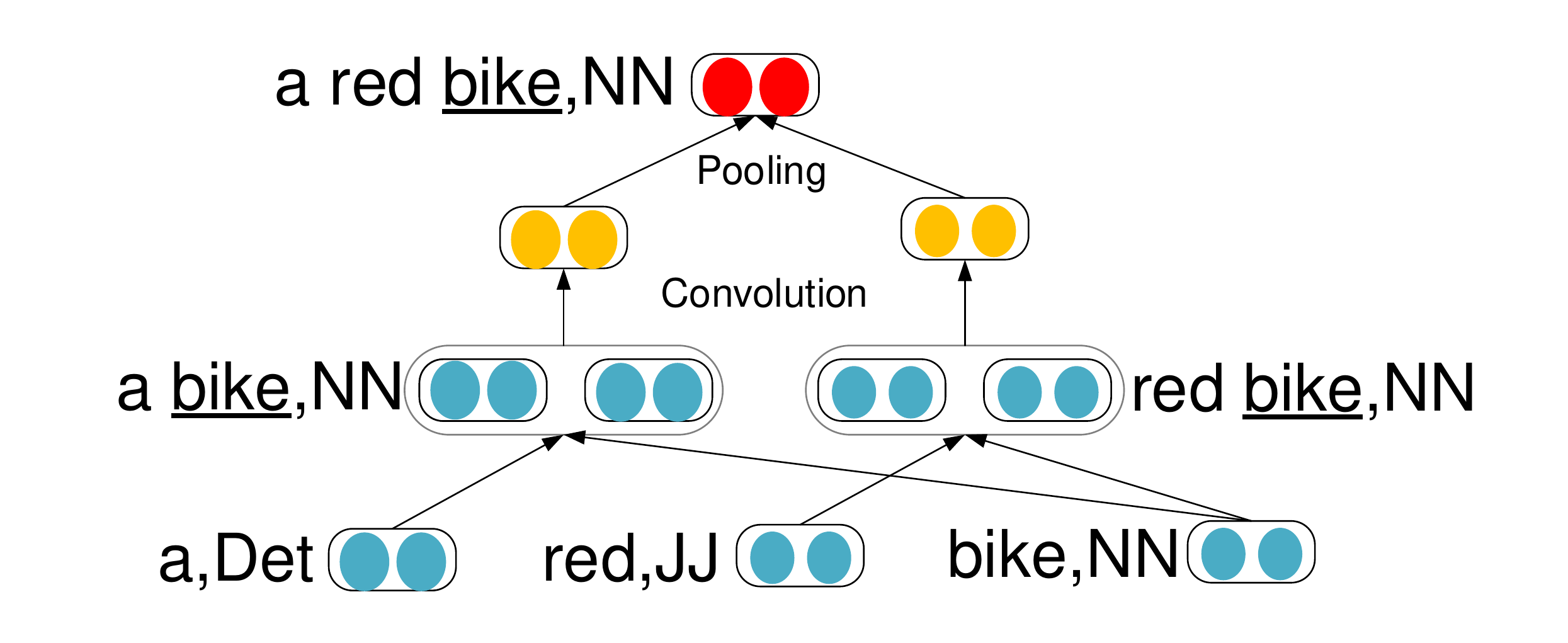}
  \caption{Illustration of a RCNN unit.}\label{fig:RCNN-Example}
\end{figure}

In this work, we address the problem to represent all level nodes (words or phrases) with dense representations in a dependency tree.
We propose a recursive convolutional neural network (RCNN) architecture to capture syntactic and compositional-semantic representations of phrases and words. RCNN is a general architecture and can deal with k-ary parsing tree, therefore it is very suitable for dependency parsing. For each node in a given dependency tree, we first use a RCNN unit to model the interactions between it and each of its children and choose the most informative features by a pooling layer. Thus, we can apply the RCNN unit recursively to get the vector representation of the whole dependency tree. The output of each RCNN unit is used as the input of the RCNN unit of its parent node,  until it outputs a single fixed-length vector at root node. Figure \ref{fig:RCNN-Example} illustrates an example how a RCNN unit represents the phrases ``a red bike'' as continuous vectors.

The contributions of this paper can be summarized as follows.
\begin{itemize}
  \item RCNN is a general architecture to model the distributed representations of a phrase or sentence with its dependency tree.  Although RCNN is just used for the re-ranking of the dependency parser in this paper, it can be regarded as semantic modelling of text sequences and handle the input sequences of varying length into a fixed-length vector. The parameters in RCNN can be learned jointly with some other NLP tasks, such as text classification.
  \item Each RCNN unit can model the complicated interactions of the head word and its children. Combined with a specific task, RCNN can capture the most useful semantic and structure information by the convolution and pooling layers.
  \item When applied to the re-ranking model for parsing, RCNN improve the accuracy of base parser to make accurate parsing decisions. The experiments on two benchmark datasets show that RCNN outperforms the state-of-the-art models.
\end{itemize}

\section{Recursive  Neural Network}

In this section, we briefly describe the recursive neural network architecture of \cite{Socher2013parsing}.

The idea of recursive neural networks (RNN) for natural language processing (NLP) is to train a deep learning model that can be applied to phrases and sentences, which have a grammatical structure \cite{pollack1990recursive,Socher:2013b}.
RNN can be also regarded as a general structure to model sentence. At every node in the tree, the contexts at the left and right children of the node are combined by a classical layer. The weights of the layer are shared across all nodes in the tree. The layer computed at the top node gives a representation for the whole sentence.

\begin{figure}
  \centering
  \includegraphics[width=0.48\textwidth]{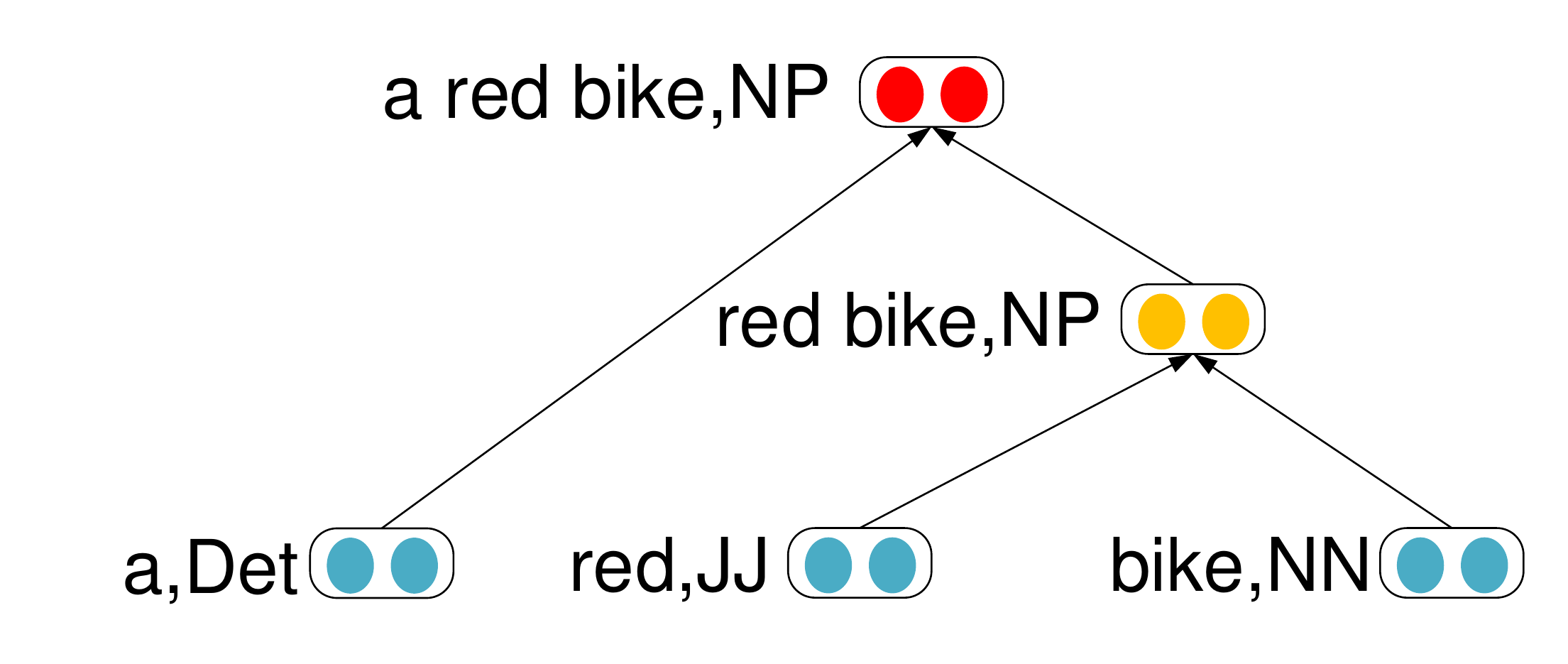}\label{fig:socherTree}
\caption{Illustration of a RNN unit.}\label{fig:rnnAndRcnn}
\end{figure}

Following the binary tree structure, RNN can assign a fixed-length vector to each word at the leaves of the tree, and combine word and phrase pairs recursively to create intermediate node vectors of the same length, eventually having one final vector representing the whole sentence. Multiple recursive combination functions have been explored, from linear transformation matrices to tensor products \cite{Socher:2013b}.
Figure \ref{fig:rnnAndRcnn} illustrates the architecture of RNN.

The binary tree can be represented in the form of branching triplets $(p \rightarrow c_1c_2)$. Each such triplet denotes that a parent node $p$ has two children and each $c_k$ can be either a word or a non-terminal node in the tree.

Given a labeled binary parse tree, $\left((p_2 \rightarrow ap_1),(p_1 \rightarrow bc)\right)$, the node representations are computed by
\begin{gather}
  \mathbf{p}_1 = f(\W\left[\begin{array}{c}
              \mathbf{b} \\
              \mathbf{c}
            \end{array}
  \right]) , \mathbf{p}_2 = f(\W\left[\begin{array}{c}
              \mathbf{a} \\
              \mathbf{p}_1
            \end{array}
  \right]),
\end{gather}
where $(\mathbf{p}_1, \mathbf{p}_2, \mathbf{a}, \mathbf{b}, \mathbf{c})$ are the vector representation of $({p}_1, {p}_2, {a}, {b}, {c})$ respectively, which are denoted by lowercase bold font letters; $\W$ is a matrix of parameters of the RNN.

Based on RNN, \newcite{Socher2013parsing} introduced a compositional vector grammar, which uses the syntactically untied weights $\W$ to learn the syntactic-semantic, compositional vector representations.
In order to compute the score of how plausible of a syntactic constituent a parent is, RNN uses a single-unit linear layer for all $p_i$:
\begin{gather}
s(p_i) = \bv \cdot \mathbf{p}_i,
\end{gather}
where $\bv$ is a vector of parameters that need to be trained. This score will be used to find the
highest scoring tree. For more details on how standard RNN can be used for parsing, see \cite{socher2011parsing}.

\newcite{costa2003towards} applied recursive neural networks to re-rank possible phrase attachments in an incremental constituency parser. Their work is the first to show that RNNs can capture enough information to make the correct parsing decisions. \newcite{menchetti2005wide} used RNNs to re-rank different constituency parses. For their results on full sentence parsing, they re-ranked candidate trees created by the Collins parser \cite{collins2003head}.

\section{Recursive Convolutional Neural Network}
The dependency grammar is a widely used syntactic structure, which directly reflects relationships among the words in a sentence.
In a dependency tree, all nodes are terminal (words) and each node may have more than two children. Therefore, the standard RNN architecture is not suitable for dependency grammar since it is based on the binary tree. In this section, we propose a more general architecture, called \textbf{recursive convolutional neural network} (RCNN), which borrows the idea of convolutional neural network (CNN) and can deal with to k-ary tree.

\subsection{RCNN Unit}
\begin{figure}
  \includegraphics[width=0.48\textwidth]{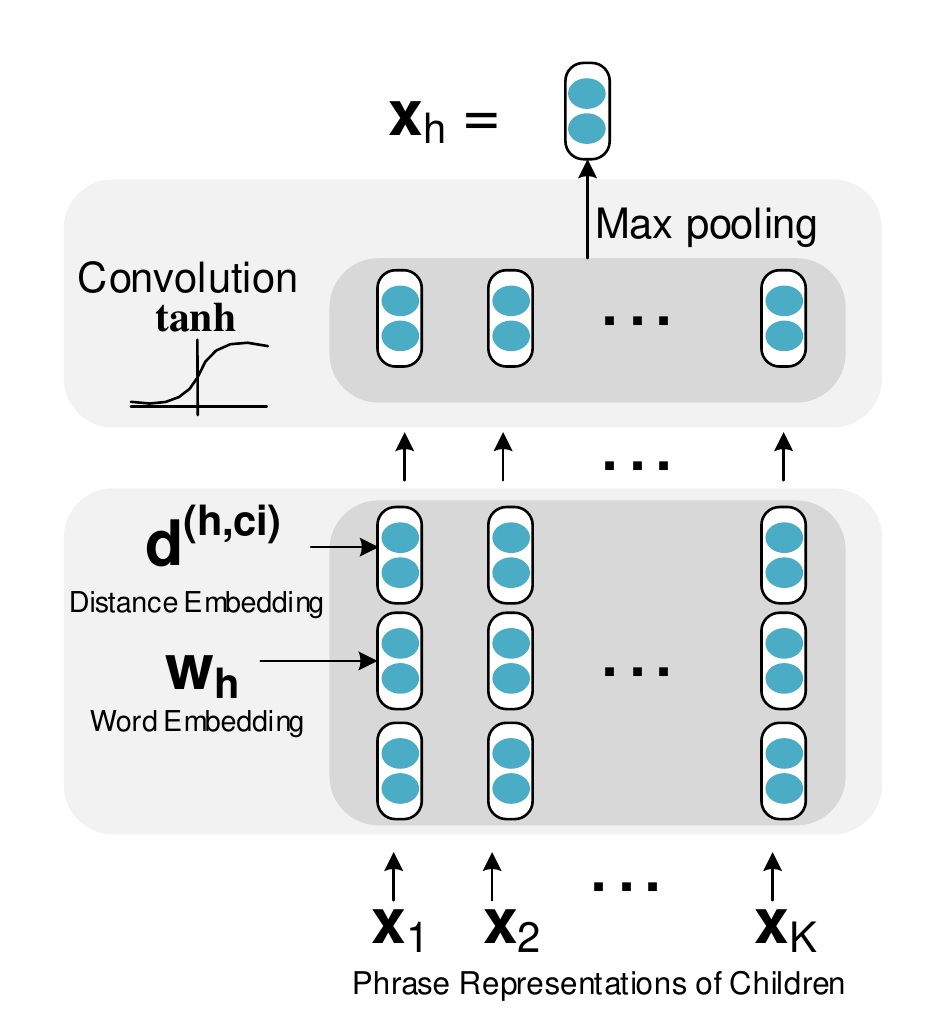}
  \hspace{1em}
  \caption{Architecture of a RCNN unit.}\label{fig:rcnn-unit}
\end{figure}

For ease of exposition, we first describe the basic unit of RCNN. A RCNN unit is to model a head word and its children.
Different from the constituent tree, the dependency tree does not have non-terminal nodes. Each node consists of  a word and its POS tags. Each node should have a different interaction with its head node.

\paragraph{Word Embeddings}
Given a word dictionary $\mathcal{W}$, each word $w \in \mathcal{W}$ is represented as a real-valued vector (word embedding) $\w \in \mathbb{R}^m$ where $m$ is the dimensionality of the vector space. The word embeddings are then stacked into a embedding matrix $M \in \mathbb{R}^{m×|\mathcal{W}|}$. For a word $w \in \mathcal{W}$, its corresponding word embedding $Embed(w) \in \mathbb{R}^m$ is retrieved by the lookup table layer. The matrix $M$ is initialized with pre-training embeddings and updated by back-propagation.

\paragraph{Distance Embeddings}

Besides word embeddings, we also use distributed vector to represent the relative distance of a head word $h$ and one of its children $c$. For example, as shown in Figure \ref{fig:RCNN-Example}, the relative distances of  ``bike''  to ``a'' and ``red'' are -2 and -1, respectively. The relative distances also are mapped to a vector of dimension $m_d$ (a hyperparameter); this vector is randomly initialized. Distance embedding is a usual way to encode the distance information in neural model, which has been proven effectively in several tasks. Our experimental results also show that the distance embedding gives more benefits than the traditional representation. The relative distance can encode the structure information of a subtree.

\paragraph{Convolution}
The word and distance embeddings are subsequently fed into the convolution component to model the interactions between two linked nodes.

Different with standard RNN, there are no non-terminal nodes in dependency tree. Each node $h$ in dependency tree has two associated distributed representations:
\begin{enumerate}
  \item word embedding $\w_h \in \mathbb{R}^m$, which is denoted as its own information according to its word form;
  \item phrase representation $\x_h \in \mathbb{R}^m$, which is denoted as the joint representation of the whole subtree rooted at $h$. In particular, when $h$ is leaf node, $\x_h =\w_h$.
\end{enumerate}

Given a subtree rooted at $h$ in dependency tree, we define $c_i, 0<i\leq L$ as the $i$-th child node of $h$, where $L$ represents the number of children.

For each pair $(h,c_i)$, we use a convolutional hidden layer to compute their combination representation $\z_i$.
 \begin{equation}
  \z_i = \tanh(\W^{(h,c_i)} \p_i), 0 < i\leq K,
\end{equation}
where $\W^{(h,c_i)} \in \mathbb{R}^{m \times n}$ is the linear \textbf{composition matrix}, which depends on the POS tags of $h$ and $c_i$; $\p_i \in \mathbb{R}^n $ is the concatenated representation of $h$ and the $i$-th child, which consists of the head word embeddings $\w_h$, the child phrase representation $\x_{c_i}$ and the distance embeddings $\bd^{h,c_i}$ of $h$ and $c_i$,
  \begin{equation}
  \p_i = \x_h \oplus  \x_{c_i}  \oplus  \bd^{(h,c_i)},
\end{equation}
where $\oplus$ represents the concatenation operation.

The distances $d_{h,c_i}$ is the relative distance of $h$ and $c_i$ in a given sentence. Then, the relative distances also are mapped to $m$-dimensional vectors. Different from constituent tree, the combination should consider the order or position of each child in dependency tree.


In our model, we do not use the POS tags embeddings directly. Since the composition matrix varies on the different pair of POS tags of $h$ and $c_i$, it can capture the different syntactic combinations. For example, the combination of adjective and noun should be different with that of verb and noun.

After the composition operations, we use \textbf{tanh} as the non-linear activation function to get a hidden representation $\z$.

\paragraph{Max Pooling}
After convolution, we get $\mathbf{Z}^{(h)} = [\z_1, \z_2, \cdots , \z_K]$, where $K$ is dynamic and depends on the number of children of $h$. To transform $\mathbf{Z}$ to a fixed length and determine the most useful semantic and structure information, we perform a max pooling operation to $\mathbf{Z}$ on rows.

\begin{equation}
  \x^{(h)}_j = \max_i \mathbf{Z}^{(h)}_{j,i}, 0 < j \leq m.
\end{equation}

Thus, we obtain the vector representation $\x_{h} \in \mathbb{R}^m$ of the whole subtree rooted at node $h$.

Figure \ref{fig:rcnn-unit} shows the architecture of our proposed RCNN unit.

Given a whole dependency tree, we can apply the RCNN unit recursively to get the vector representation of the whole sentence. The output of each RCNN unit is used as the input of the RCNN unit of its parent node.

Thus, RCNN can be used to model the distributed representations of a phrase or sentence with its dependency tree and applied to many NLP tasks. The parameters in RCNN can be learned jointly with the specific NLP tasks.
Each RCNN unit can model the complicated interactions of the head word and its children. Combined with a specific task, RCNN can select the useful semantic and structure information by the convolution and max pooling layers.

\begin{figure}
  \centering
  \includegraphics[width=0.5\textwidth]{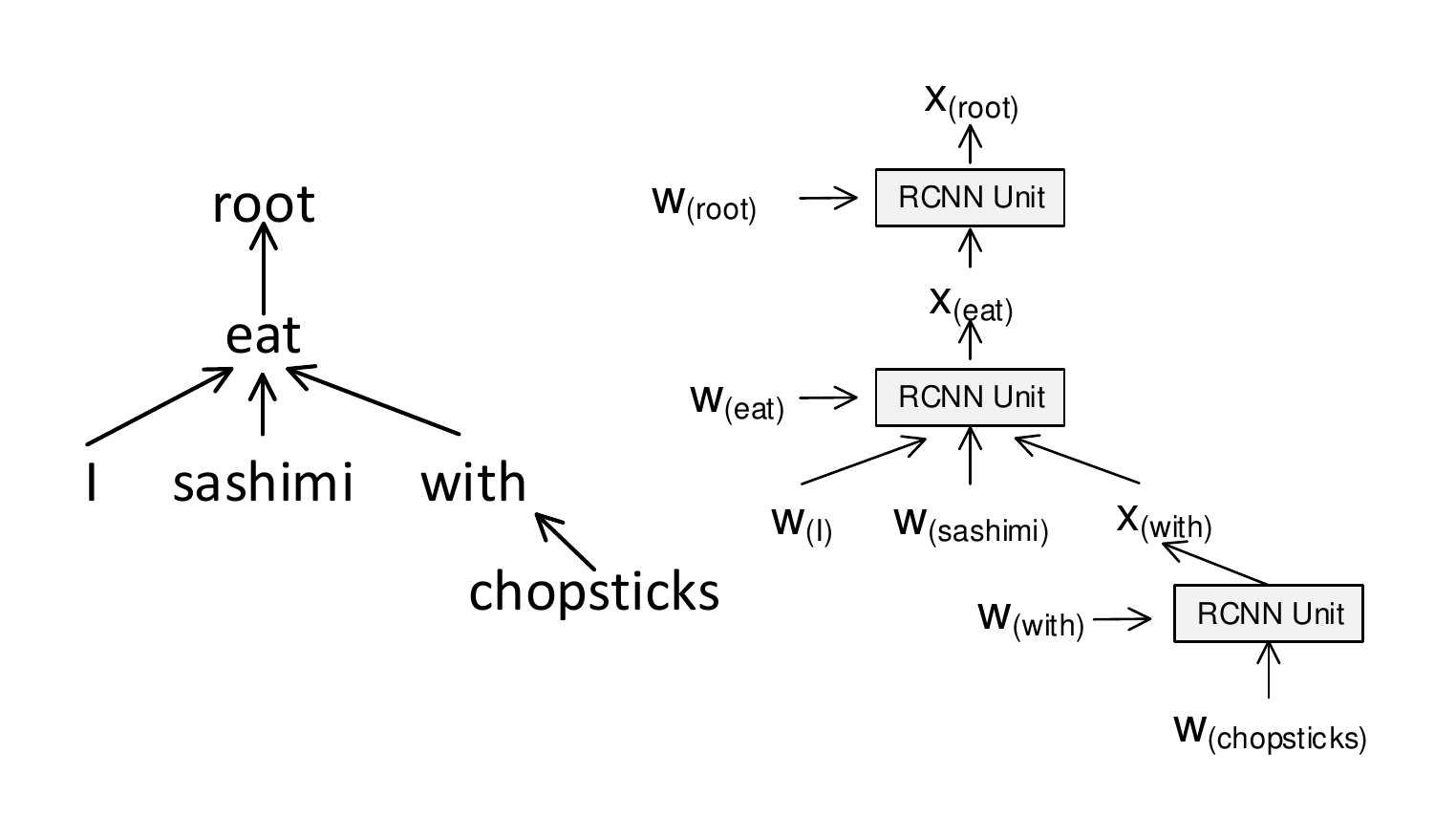}
  \caption{Example of a RCNN unit}\label{fig:RCNN-Tree}
\end{figure}

Figure \ref{fig:RCNN-Tree} shows an example of RCNN to model the sentence ``\textit{I eat sashimi with chopsitcks}''.

\section{Parsing}

In order to measure the plausibility of a subtree rooted at $h$ in dependency tree, we use a single-unit linear layer neural network to compute the score of its RCNN unit.

For constituent parsing, the representation of a non-terminal node only depends on its two children. The combination is relative simple and its correctness can be measured with the final representation of the non-terminal node \cite{Socher2013parsing}.

However for dependency parsing, all combinations of the head $h$ and its children $c_i(0 < i\leq K)$ are important to measure the correctness of the subtree. Therefore, our score function $s(h)$ is computed on all of hidden layers $\z_i(0 < i\leq K)$:
\begin{equation}
  s(h) = \sum_{i=1}^K \bv^{(h,c_i)} \cdot \z_i,
\end{equation}
where $\bv^{(h,c_i)} \in \mathbb{R}^{m \times 1}$ is the \textbf{score vector}, which also depends on the POS tags of $h$ and $c_i$.

Given a sentence $x$ and its dependency tree $y$, the goodness of a complete tree is measured by summing the scores of all the RCNN units.
\begin{equation}
  s(x, y, \Theta) = \sum_{h \in y} s(h),
\end{equation}
where $h\in y$ is the node in tree $y$; $\Theta=\{\Theta_\W, \Theta_\bv, \Theta_{\w}, \Theta_{\bd} \}$ including the combination matrix set $\Theta_\W$, the score vector set $\Theta_\bv$, the word embeddings $\Theta_{\w}$ and distance embeddings $\Theta_{\bd}$.

Finally, we can predict dependency tree $\hat{y}$ with highest score for sentence $x$.
\begin{equation}
  \hat{y} = \argmax_{y \in \textbf{gen}(x)} s(x, y, \Theta), \label{eq:parsing}
\end{equation}
where $\textbf{gen}(x)$ is defined as the set of all possible trees for sentence $x$. When applied in re-ranking, $\textbf{gen}(x)$ is the set of the $k$-best outputs of a base parser.

\section{Training}

For a given training instance $(x_i, y_i)$, we use the max-margin criterion to train our model. We first predict the dependency tree $\hat{y_i}$ with the highest score for each $x_i$ and define a structured margin loss $\Delta(y_i, \hat{y_i})$ between the predicted tree $\hat{y_i}$ and the given correct tree $y_i$. $\Delta(y_i, \hat{y_i})$ is measured by counting the number of nodes $y_i$ with an incorrect span (or label) in the proposed tree \cite{goodman1998parsing}.
\begin{equation}
  \Delta(y_i, \hat{y_i}) = \sum_{d\in \hat{y_i}} \kappa \textbf{1}\{d\notin y_i\}
\end{equation}
where $\kappa$ is a discount parameter and $d$ represents the nodes in trees.

Given a set of training dependency parses $\mathcal{D}$, the final training objective is to minimize the loss function $J(\Theta)$, plus a $l_2$-regulation term:
\begin{equation}
  J(\Theta) = \frac{1}{|\mathcal{D}|} \sum_{(x_i,y_i)\in \mathcal{D}} r_i(\Theta) + \frac{\lambda}{2}\|\Theta\|_2^2,
\end{equation}
where
\begin{multline}
  r_i(\Theta) = \max_{\hat{y_i}\in Y(x_i)}\left(\right.0, s_t(x_i, \hat{y_i}, \Theta)\\
  +\Delta(y_i, \hat{y_i})  -s_t(x_i, y_i, \Theta)\left.\right).
\end{multline}

By minimizing this object, the score of the correct tree $y_i$ is increased and the score of the highest scoring incorrect tree
$\hat{y_i}$ is decreased.

We use a generalization of gradient descent called subgradient method \cite{ratliff:2007} which computes a gradient-like direction. The subgradient of equation is:
\begin{gather}
  \frac{\partial J}{\partial \Theta} = \frac{1}{|\mathcal{D}|}\sum_{(x_i,y_i)\in \mathcal{D}}(\frac{\partial s_t(x_i, \hat{y}_i, \Theta)}{\partial \Theta} -\nonumber\\
  \frac{\partial s_t(x_i, y_i, \Theta)}{\partial \Theta}) + \lambda\Theta.
\end{gather}

To minimize the objective, we use the diagonal variant of AdaGrad \cite{duchi2011adaptive}. The parameter update for the $i$-th parameter $\Theta_{t, i}$ at time step $t$ is as follows:
\begin{equation}
  \Theta_{t, i} = \Theta_{t-1, i} - \frac{\rho}{\sqrt{\sum\nolimits_{\tau=1}^{t}g_{\tau,i}^2}}g_{t, i},
\end{equation}
where $\rho$ is the initial learning rate and $g_\tau \in \mathbb{R}^{|\theta_i|}$ is the subgradient at time step $\tau$ for parameter $\theta_{i}$.

\section{Re-rankers}

Re-ranking $k$-best lists was introduced by \newcite{Collins:2005} and \newcite{Charniak:2005}. They used discriminative methods to re-rank the constituent parsing. In the dependency parsing, \newcite{Sangati:2009} used a third-order generative model for re-ranking $k$-best lists of base parser. \newcite{Hayashi:2013} used a discriminative forest re-ranking algorithm for dependency parsing. These re-ranking models achieved a substantial raise on the parsing performances.

Given $\mathcal{T}(x)$, the set of $k$-best trees of a sentence $x$ from a base parser, we use the popular mixture re-ranking strategy \cite{Hayashi:2013,le2014distributed}, which is a combination of the our model and the base parser.
\begin{equation}
  \hat{y_i} = \argmax_{y \in \mathcal{T}(x_i)} \alpha s_t(x_i, y, \Theta) + (1-\alpha) s_b(x_i,y)\label{eq:re-ranker}
\end{equation}
where $\alpha \in [0, 1]$ is a hyperparameter; $s_t(x_i, y, \Theta)$ and $s_{b}(x_i,y)$ are the scores given by RCNN and the base parser respectively.

To apply RCNN into re-ranking model, we first get the $k$-best outputs of all sentences in train set with a base parser. Thus, we can train the RCNN in a discriminative way and optimize the re-ranking strategy for a particular base parser.

Note that the role of RCNN is not fully valued when applied in re-ranking model since that the $\textbf{gen}(x)$ in Eq.(\ref{eq:parsing}) is just the $k$-best outputs of a base parser, not the set of all possible trees for sentence $x$. The parameters of RCNN could overfit to $k$-best outputs of training set.

\section{Experiments}\label{sec:experiments}

\subsection{Datasets}
To empirically demonstrate the effectiveness of our approach, we use two datasets in different languages (English and Chinese) in our experimental evaluation and compare our model against the other state-of-the-art methods using the unlabeled attachment score (UAS) metric ignoring punctuation.

\begin{description}
  \item[English] For English dataset, we follow the standard splits of Penn Treebank (PTB), using sections 2-21 for training, section 22 as development set and section 23 as test set. We tag the development and test sets using an automatic POS tagger (at 97.2\% accuracy), and tag the training set using four-way jackknifing similar to \cite{Collins:2005}.
  \item[Chinese] For Chinese dataset, we follow the same split of the Penn Chinese Treeban (CTB5) as described in \cite{Zhang:2008} and use sections 001-815, 1001-1136 as training set, sections 886-931, 1148- 1151 as development set, and sections 816-885, 1137-1147 as test set. Dependencies are converted by using the Penn2Malt tool with the head-finding rules of \cite{Zhang:2008}. And following \cite{Zhang:2008} \cite{Zhang:2011}, we use gold segmentation and POS tags for the input.
\end{description}

We use the linear-time incremental parser \cite{Huang:2010} as our base parser and calculate the $64$-best parses at the top cell of the chart. Note that we optimize the training settings for base parser and the results are slightly improved on \cite{Huang:2010}. Then we use max-margin criterion to train RCNN. Finally, we use the mixture strategy to re-rank the top $64$-best parses.

For initialization of parameters, we train word2vec embeddings \cite{Mikolov:2013} on Wikipedia corpus for English and Chinese respectively.
For the combination matrices and score vectors, we use the random initialization within $(−0.01, 0.01)$. The parameters which achieve the best unlabeled attachment score on the development set will be chosen for the final evaluation.

\subsection{English Dataset}

\begin{figure*}[t]
  \centering
  \pgfplotsset{width=0.48\textwidth}
  \begin{tikzpicture}
    \begin{axis}[
    xlabel={$k$},
    xtick=data,
    xticklabels={1,2,3,4,5,6,7,8,9,10,32,64},
    ylabel={UAS(\%)},
    legend entries={Oracle Best,RCNN,Re-ranker},
    legend pos= north west,
    ymajorgrids=true,
    grid style=dashed,
    ]
    \addplot table [x index=0, y index=2] {rerank-dev.txt};
    \addplot table [x index=0, y index=5] {rerank-dev.txt};
    \addplot table [x index=0, y index=6] {rerank-dev.txt};
    \end{axis}
\end{tikzpicture} 
\pgfplotsset{width=0.48\textwidth}
  \begin{tikzpicture}
    \begin{axis}[
    xlabel={$k$},
    xtick=data,
    xticklabels={1,2,3,4,5,6,7,8,9,10,32,64},
    ylabel={UAS(\%)},
    legend entries={Oracle Best,RCNN,Re-ranker, Oracle Worst},
    legend pos= south west,
    ymajorgrids=true,
    grid style=dashed,
    ]
    \addplot table [x index=0, y index=2] {rerank-dev.txt};
    \addplot table [x index=0, y index=5] {rerank-dev.txt};
    \addplot table [x index=0, y index=6] {rerank-dev.txt};
    \addplot table [x index=0, y index=3] {rerank-dev.txt};
    \end{axis}
\end{tikzpicture}\\
(a) without the oracle worst result \hspace{3cm} (b) with the oracle worst result

  \caption{UAS with varying $k$ on the development set. Oracle best: always choosing the best result in the $k$-best of base parser; Oracle worst: always choosing the worst result in the $k$-best of base parser; RCNN: choosing the most probable candidate according to the score of RCNN; Re-ranker: a combination of the RCNN and base parser.}\label{fig:rerank-ptb-dev}
\end{figure*}
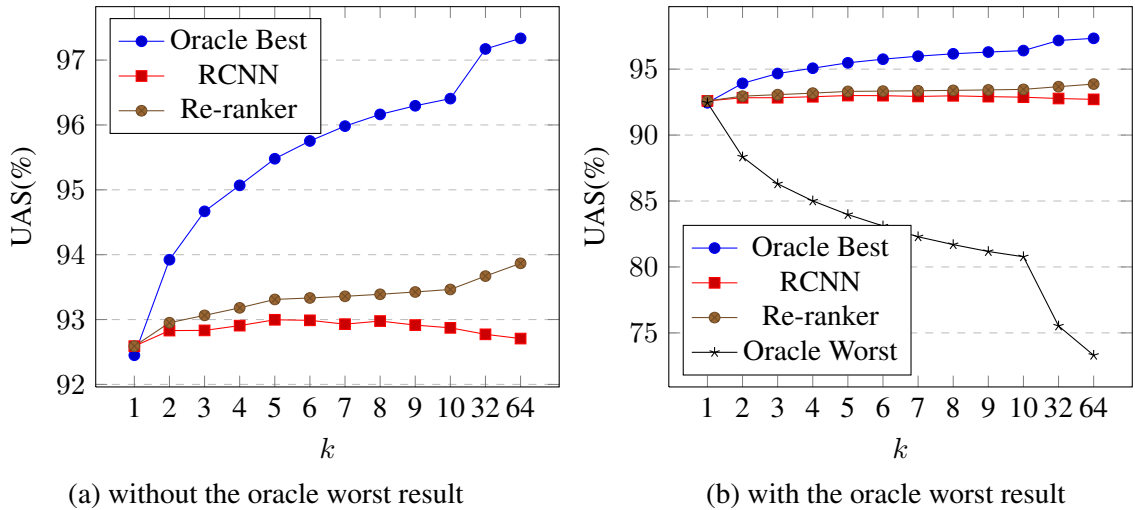

We first evaluate the performances of the RCNN and re-ranker (Eq. (\ref{eq:re-ranker})) on the development set. Figure \ref{fig:rerank-ptb-dev} shows UASs of different models with varying $k$. The base parser achieves 92.45\%. When $k = 64$, the oracle best of base parser achieves 97.34\%, while the oracle worst achieves 73.30\% (-19.15\%) . RCNN achieves the maximum improvement of 93.00\%(+0.55\%) when $k = 6$. When $k>6$, the performance of RCNN declines with the increase of $k$ but is still higher than baseline (92.45\%). The reason behind this is that RCNN could require more negative samples to  avoid overfitting when $k$ is large. Since the negative samples are limited in the $k$-best outputs of a base parser, the learnt parameters could easily overfits to the training set.

The mixture re-ranker achieves the maximum improvement of 93.50\%(+1.05\%) when $k = 64$. In mixture re-ranker, $\alpha$ is optimised by searching with the step-size 0.005.

Therefore, we use the mixture re-ranker in the following experiments since it can take the advantages of both the RCNN and base models.

\begin{figure}[t]
  \centering
\pgfplotsset{width=0.48\textwidth}
\begin{tikzpicture}
 \pgfplotstableread{pos-dev.txt}\loadedtable;
    \begin{axis}[ybar,bar width=4pt,
     symbolic x coords={WRB,JJR,WDT,VBG,VBP,JJS,RB,CC,MD,IN},
     ylabel={UAS(\%)},
     ymajorgrids=true,
       tickwidth=0pt,
        enlarge x limits=true,
        xticklabel style={font=\tiny},
        xtick=data,
        legend style={
            at={(0.5,-0.1)},
            anchor=north,
            legend columns=-1,
            /tikz/every even column/.append style={column sep=0.5cm}
        },
     ]
        \addplot [cyan!50!black] table [x=POS,y=Base] {\loadedtable};
        \addplot table [x=POS,y=Ours] {\loadedtable};
        \legend{Base Paser, Re-ranker}
    \end{axis}
\end{tikzpicture}
  \caption{Accuracies on the top ten POS tags of the modifier words with the largest improvements on the development set.}\label{fig:res-pos}
\end{figure}
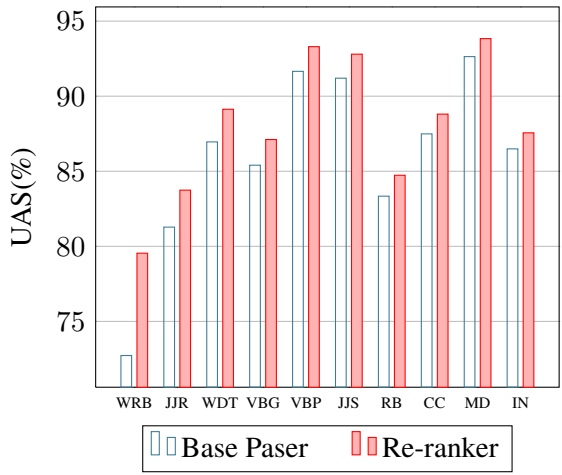

Figure \ref{fig:res-pos} shows the accuracies on the top ten POS tags of the modifier words with the largest improvements. We can see that our re-ranker can improve the accuracies of \textit{CC} and \textit{IN}, and therefore may indirectly result in rising the the well-known coordinating conjunction and PP-attachment problems.

The final experimental results on test set are shown in Table \ref{tab:res-eng}. The hyperparameters of our model are set as in Table \ref{tab:hyper}. Our re-ranker achieves the maximum improvement of 93.83\%(+1.48\%) on test set.  Our system performs slightly better than many state-of-the-art systems such as \newcite{Zhang:2008} and \newcite{Huang:2010}. It outperforms \newcite{Hayashi:2013} and \newcite{le2014inside}, which also use the mixture re-ranking strategy.

Since the result of ranker is conditioned to $k$-best results of base parser, we also do an experiment to avoid this limitation by adding the oracle to $k$-best candidates. With including oracle, the re-ranker can achieve 94.16\% on UAS, which is shown in the last line (``our re-ranker (with oracle)'') of Table \ref{tab:res-eng}.

\begin{table}[!t]
\centering
\begin{tabular}{l|c}
  \hline
   & UAS\\
   \hline
   \textbf{Traditional Methods}\\
    \newcite{Zhang:2008} &  91.4 \\
    \newcite{Huang:2010} &  92.1 \\
    \hline
    \textbf{Distributed Representations}\\
    \newcite{Stenetorp:2013} &  86.25  \\
    \newcite{Chen:2014} &   93.74 \\
    \newcite{Chen:2014a} &  92.0   \\
    \hline
    \textbf{Re-rankers}&\\
    \newcite{Hayashi:2013} & 93.12 \\
    \newcite{le2014inside} & 93.12  \\
    \hline
    Our baseline & 92.35 \\
    Our re-ranker &  93.83(+1.48)\\
    Our re-ranker (with oracle) & 94.16 \\
   \hline
\end{tabular}
\caption{Accuracy on English test set. Our baseline is the result of base parser; our re-ranker uses the mixture strategy on the 64-best outputs of base parser; our re-ranker(with oracle) is to add the oracle to $k$-best outputs of base parser.}\label{tab:res-eng}
\end{table}

\begin{table}
\centering
\begin{tabular}{l|c}
  \hline
   Word embedding size & $m=25$ \\
   Distance embedding size & $m_d=25$ \\
   Initial learning rate & $\rho=0.1$\\
   Margin loss discount & $\kappa=2.0$\\
   Regularization & $\lambda=10^{-4}$\\
   $k$-best & $k = 64$\\
  \hline
\end{tabular}
\caption{Hyperparameters of our model}\label{tab:hyper}
\end{table}


\subsection{Chinese Dataset}
We also make experiments on the Penn Chinese Treebank (CTB5). The hyperparameters is the same as the previous experiment on English except that $\alpha$ is optimised by searching with the step-size 0.005.

The final experimental results on the test set are shown in Table \ref{tab:res-chn}. Our re-ranker achieves the performance of 85.71\%(+0.25\%) on the test set, which also outperforms the previous state-of-the-art methods. With adding oracle, the re-ranker can achieve 87.43\% on UAS, which is shown in the last line (``our re-ranker (with oracle)'') of Table \ref{tab:res-chn}. Compared with the re-ranking model of \newcite{Hayashi:2013}, that use a large number of handcrafted features, our model can achieve a competitive performance with the minimal feature engineering.

\begin{table}
\centering
\begin{tabular}{l|c}
  \hline
   & UAS\\
   \hline
   \textbf{Traditional Methods}\\
    \newcite{Zhang:2008}   & 84.33 \\
    \newcite{Huang:2010} & 85.20 \\
    \hline
    \textbf{Distributed Representations}\\
    \newcite{Chen:2014}   & 82.94 \\
    \newcite{Chen:2014a}  & 83.9   \\
    \hline
    \textbf{Re-rankers}\\
    \newcite{Hayashi:2013} & 85.9 \\
    \hline
    Our baseline &  85.46  \\
    Our re-ranker & \textbf{85.71}(+0.25)\\
    Our re-ranker (with oracle) & 87.43 \\
   \hline
\end{tabular}
\caption{Accuracy on Chinese test set.}\label{tab:res-chn}
\end{table}

\subsection{Discussions}

The performance of the re-ranking model is affected by the base parser. The small divergence of the dependency trees in the output list also results to overfitting in training phase. Although our re-ranker outperforms the state-of-the-art methods, it can also benefit from improving the quality of the candidate results. It was also reported in other re-ranking works that a larger $k$ (eg. $k > 64$) results the worse performance. We think the reason is that the oracle best increases when $k$ is larger, but the oracle worst decrease with larger degree. The error types increase greatly. The re-ranking model requires more negative samples to avoid overfitting. When $k$ is larger, the number of negative samples also needs to multiply increase for training. However, we just can obtain at most $k$ negative samples from the k-best outputs of the base parser.

 The experiments also show that the our model can achieves significant improvements by adding the oracles into the output lists of the base parser. This indicates that our model can be boosted by a better set of the candidate results, which can be implemented by combining the RCNN in the decoding algorithm.

\section{Related Work}\label{sec:related}

There have been several works to use neural networks and distributed representation for dependency parsing.

\newcite{Stenetorp:2013} attempted to build recursive neural networks for transition-based dependency parsing, however the empirical performance of his model is still unsatisfactory. \newcite{Chen:2014a} improved the transition-based dependency parsing by representing all words, POS tags and arc labels as dense vectors, and modeled their interactions with neural network to make predictions of actions.
Their methods aim to transition-based parsing and can not model the sentence in semantic vector space for other NLP tasks.

\begin{figure}
  \centering
  \includegraphics[width=0.48\textwidth]{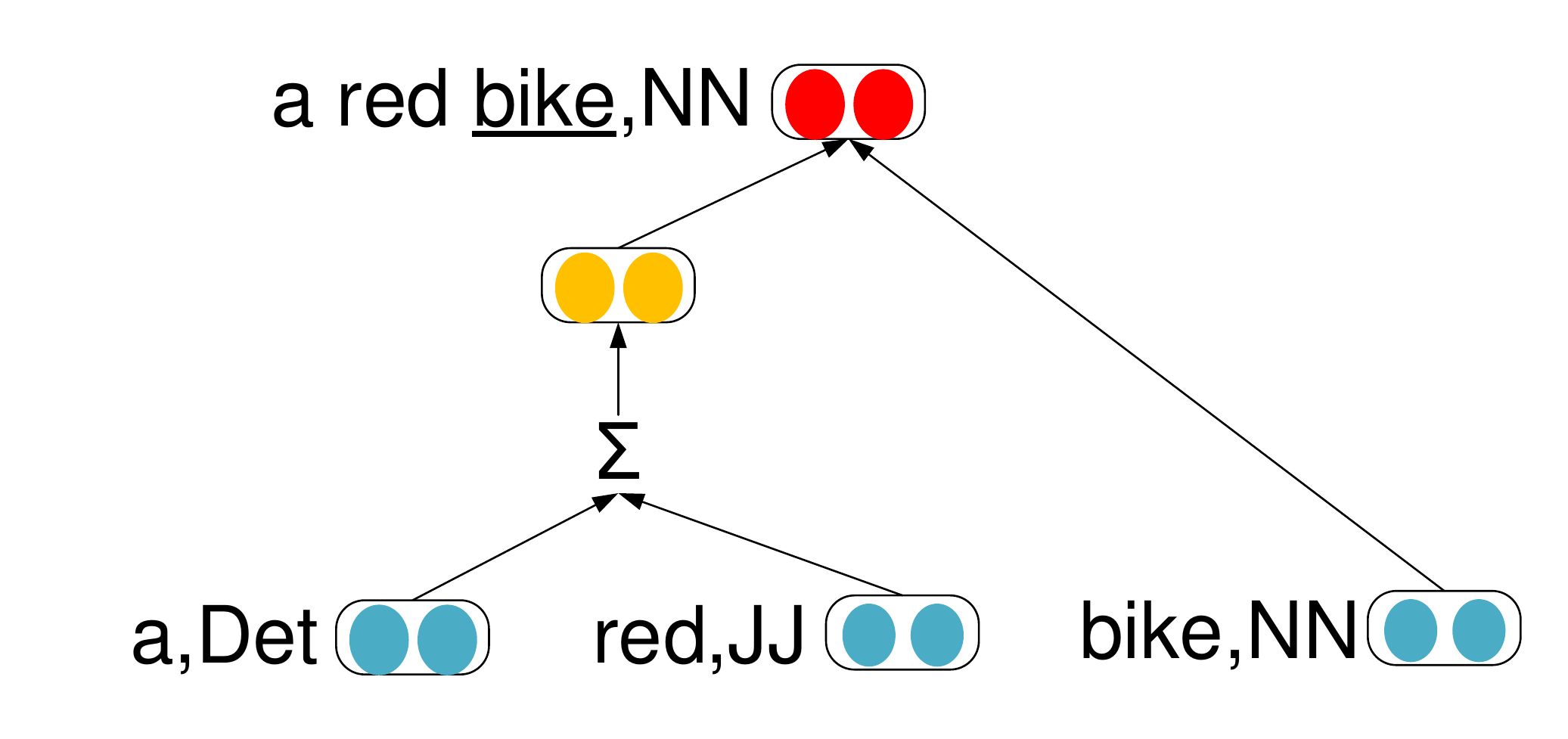}
  \caption{Example of a DT-RNN unit}\label{fig:DTRNN}
\end{figure}

\newcite{socher2013grounded} proposed a compositional vectors computed by dependency tree RNN (DT-RNN) to map sentences and images
into a common embedding space. However, there are two major differences as follows. 1) They first summed up all child nodes into a dense vector $\bv_c$ and then composed subtree representation from $\bv_c$ and vector parent node. In contrast, our model first combine the parent and each child and then choose the most informative features with a pooling layer.
2) We represent the relative position of each child and its parent with distributed representation (position embeddings), which is very useful for convolutional layer. Figure \ref{fig:DTRNN} shows an example of DTRNN to illustrates how RCNN represents phrases as continuous vectors.

Specific to the re-ranking model, \newcite{le2014inside} proposed a generative re-ranking model with Inside-Outside Recursive Neural Network (IORNN), which can process trees both bottom-up and top-down. However, IORNN works in generative way and just estimates the probability of a given tree, so IORNN cannot fully utilize the incorrect trees in $k$-best candidate results. Besides, IORNN treats dependency tree as a sequence, which can be regarded as a generalization of simple recurrent neural network (SRNN) \cite{Elman:1990}. Unlike IORNN, our proposed RCNN is a discriminative model and can optimize the re-ranking strategy for a particular base parser. Another difference is that RCNN computes the score of tree in a recursive way, which is more natural for the hierarchical structure of natural language. Besides, the RCNN can not only be used for the re-ranking, but also be regarded as general model to represent sentence with its dependency tree.

\section{Conclusion}\label{sec:conclusion}

In this work, we address the problem to represent all level nodes (words or phrases) with dense representations in a dependency tree.
We propose a recursive convolutional neural network (RCNN) architecture to capture the syntactic and compositional-semantic representations of phrases and words. RCNN is a general architecture and can deal with k-ary parsing tree, therefore RCNN is very suitable for many NLP tasks to minimize the effort in feature engineering with a external dependency parser. 
Although RCNN is just used for the re-ranking of the dependency parser in this paper, it can be regarded as semantic modelling of text sequences and handle the input sequences of varying length into a fixed-length vector. The parameters in RCNN can be learned jointly with some other NLP tasks, such as text classification.

For the future research, we will develop an integrated parser to combine RCNN with a decoding algorithm. We believe that the integrated parser can achieve better performance without the limitation of base parser. Moreover, we also wish to investigate the ability of our model for other NLP tasks.

\section*{Acknowledgments}
We would like to thank the anonymous reviewers for their valuable comments. This work was partially funded by National Natural Science Foundation of China (61472088, 61473092), The National High Technology Research and Development Program of China (2015AA015408), Shanghai Science and Technology Development Funds (14ZR1403200), Shanghai Leading Academic Discipline Project (B114).

\bibliographystyle{acl}

\bibliography{../nlp}

\end{document}